\documentclass{article}
\usepackage{arxiv}
\usepackage{times}
\usepackage{soul}
\usepackage[utf8]{inputenc}
\usepackage{graphicx}
\usepackage{booktabs}
\usepackage{algorithm}
\usepackage[noend]{algorithmic}

\usepackage{helvet}
\usepackage{courier}
\usepackage[hyphens]{url}
\usepackage{natbib}

\usepackage{microtype}
\usepackage{xcolor}

\newcommand{\bigCI}{\mathrel{\text{\scalebox{1.07}{$\perp\mkern-10mu\perp$}}}}
\usepackage{makecell}
\usepackage{multirow}
\usepackage{import}
\usepackage{tikz}   
\usetikzlibrary{matrix,chains,positioning,decorations.pathreplacing,arrows}
\usepackage{epstopdf}

\usepackage{amsmath,amssymb,centernot}

\DeclareMathOperator*{\glm}{glm}

\newcommand{\V}{\mathbf{V}}
\newcommand{\X}{\mathbf{X}}
\newcommand{\Y}{\mathbf{Y}}
\newcommand{\Z}{\mathbf{Z}}
\newcommand{\x}{\mathbf{x}}

\newcommand{\cbar}{\,|\,}

\usepackage[amsmath,thmmarks]{ntheorem}
\theoremseparator{.}
\theoremstyle{change} 

\renewtheorem{theorem*}{Theorem}

\renewtheorem{definition*}{Definition}

\title{Conditional Sum-Product Networks:\\ Imposing Structure on Deep Probabilistic Architectures}

\author{
\Large \textbf{Xiaoting Shao\textsuperscript{\rm 1}, Alejandro Molina\textsuperscript{\rm 1}, Antonio Vergari\textsuperscript{\rm 3}} \\
\Large \textbf{Karl Stelzner\textsuperscript{\rm 1}, Robert Peharz\textsuperscript{\rm 4}, Thomas Liebig\textsuperscript{\rm 5}, Kristian Kersting\textsuperscript{\rm 1, \rm 2}} \\
\textsuperscript{\rm 1}CS Department, and \textsuperscript{\rm 2}Centre for Cognitive Science, TU Darmstadt, Germany, 
\textsuperscript{\rm 3}CS Department, UCLA, USA \\
\textsuperscript{\rm 4}CBL Lab, University of Cambridge, UK, 
\textsuperscript{\rm 5}CS Department, TU Dortmund, Germany
}

\begin{document}
\maketitle

\begin{abstract}
Probabilistic graphical models are a central tool in AI; however, they are generally
not as expressive as deep neural models, and inference is notoriously hard and slow.
In contrast, deep probabilistic models such as sum-product networks (SPNs) capture
joint distributions in a tractable fashion, but still lack the expressive power of
intractable models based on deep neural networks.
Therefore, we introduce conditional SPNs (CSPNs), conditional density estimators for multivariate and 
potentially hybrid domains which allow harnessing the expressive power of neural networks
while still maintaining tractability guarantees.
One way to implement CSPNs is to use an existing SPN structure and condition its parameters on the input,
e.g., via a deep neural network.
This approach, however, might misrepresent
the conditional independence structure present in data. 
Consequently, we also develop a structure-learning approach that derives both the
structure and parameters of CSPNs from data.
Our experimental evidence demonstrates that CSPNs are competitive with other probabilistic models and
yield superior performance on multilabel image classification compared to mean field and mixture
density networks. Furthermore, they can successfully be employed as building blocks for
structured probabilistic models, such as autoregressive image models.
\end{abstract}

\section{Introduction}

Probabilistic models are a fundamental approach in machine learning and artificial intelligence to represent and distill meaningful representations from data with inherent structure. 
In practice, however, it has been challenging to come up with probabilistic models that balance three desirable goals: 1) being {expressive} enough to capture the complexity of real-world distributions; 2) maintain---at least on a high level---interpretable domain {structure}; and 3) permit a rich set of {tractable inference} routines.

Probabilistic graphical models (PGMs), for example, admit an interpretable structure, but are known to achieve a bad trade-off between expressivity and tractable inference \citep{koller2009}.
Probabilistic models based on deep neural networks such as variational autoencoders (VAEs) \citep{Kingma2014} and generative adversarial networks (GANs) \citep{Goodfellow2014} are highly expressive,
but generally lack interpretable structure and have limited capabilities when it comes
to probabilistic inference. Meanwhile, advances in deep probabilistic learning have
shown that tractable models like arithmetic circuits \citep{Darwiche2003}
and sum-product networks \citep{Poon2011}
can be used to capture
complex distributions, while maintaining a rich set of tractable inference routines.

\begin{figure}[t!]
\centering
\begin{minipage}[b]{0.45\columnwidth}
\centering
\includegraphics[scale=0.4]{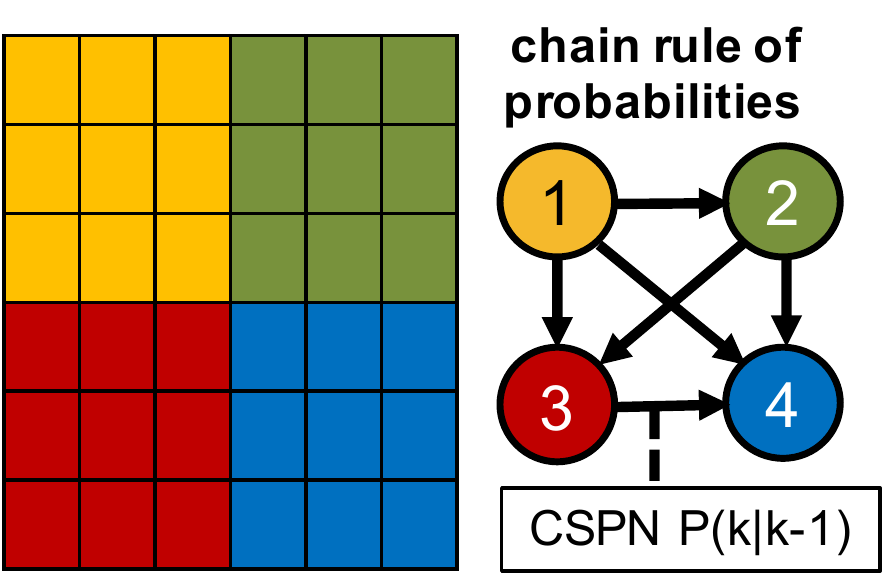}
\end{minipage}
\ \
\begin{minipage}[b]{0.45\columnwidth}
\centering
\includegraphics[scale=0.5]{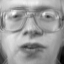}\hspace{-3pt}
\includegraphics[scale=0.5]{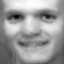}\hspace{-3pt}
\includegraphics[scale=0.5]{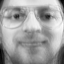}\hspace{-3pt}
\includegraphics[scale=0.5]{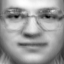}\\
\includegraphics[scale=0.5]{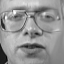}\hspace{-3pt}
\includegraphics[scale=0.5]{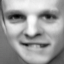}\hspace{-3pt}
\includegraphics[scale=0.5]{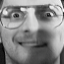}\hspace{-3pt}
\includegraphics[scale=0.5]{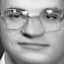}
\end{minipage}
\caption{Imposing structure on deep probabilitic architectures. (Left) An Autoregressive Block-wise CSPN (ABCSPN) factorizes a distribution over images along image patches. (Right) Samples generated by an ABCSPN (top) and a PixelCNN++ (bottom) trained on Olivetti faces.\label{fig:pixelolivetti}}
\end{figure}

How can we combine these lines of research, in order to strive for a sensible balance between expressivity, structure, and tractable inference?
This question has received surprisingly little attention, yet. 
Johnson {\it et al.}~(\citeyear{Johnson2016b}) combine VAEs with PGMs, and thus effectively equip expressive neural-based models with a rich structure.
However, inference remains intractable in these models.
Shen {\it et al.}~(\citeyear{Shen2019}) proposed structured Bayesian networks, which impose structure among clusters of variables using probabilistic sentential decision diagrams (PSDDs) to model high-dimensional conditional distributions \citep{Shen2018}. While this yields well-structured models with a wide range of tractable inference scenarios, so far they are restricted to binary data and do not establish a link to deep neural networks. Moreover, PSDDs are a more restricted than SPNs. Recently and independent of us, Rahman {\it et al.}~(\citeyear{rahman19}) proposed conditional cutset networks, which are a strict sub-class of SPNs, do not employ conditional independency tests for learning, nor conditioning on deep neural networks.

In this paper, we introduce conditional sum-product networks (CSPNs), a conditional variant of SPNs which can harness the expressive power of universal function approximators such as neural networks, while still maintaining a wide range of inference routines.
Moreover, CSPNs can be naturally used as building blocks in order to impose structure on probabilistic deep learning models.
We formally define CSPNs, provide a learning framework for them, and provide arguments for why
CSPNs are more compact than SPNs.

We also consider neural CSPNs, which rely on random SPN structures parameterized
by the output of deep neural networks. While this approaches does not have the benefit of
carefully learned structures, it gains expressiveness through increased model size.

Finally, we note that CSPNs can be naturally combined with other (C)SPNs to
impose a rich structure on high-dimensional joint distributions.
We illustrate this by introducing autoregressive 
SPNs for conditional image generation, modelling images block by block,
decomposing the joint image
distribution into a product of (C)SPNs, cf.~Fig.~\ref{fig:pixelolivetti}.

To summarize, we make the following contributions: 

\begin{itemize}
    \item 
We introduce CSPNs as a deep tractable model for modelling
multivariate, conditional probability distributions
$P(\Y \cbar \X)$ over mixed variables $\Y$, by introducing gating nodes as mixtures with
 functional mixing coefficients.  
 \item
We present a structure learning algorithm for CSPNs based on randomized conditional
correlation tests (RCoT) which
allows to learn structures from heterogeneous data sources.
 \item
We connect CSPNs with deep neural networks, and demonstrate that the improved ability of the resulting
neural CSPNs to model dependencies results in increased multilabel image
classification performance.
 \item
We illustrate how CSPNs may be used to build structured probabilistic models by 
introducing ABCSPNs, autoregressive image models based on CSPNs.
\end{itemize}

We proceed as follows. We start off by discussing related work in more detail. 
Then we introduce CSPNs, show how to learn their structure 
using RCoT, and devise autoregressive SPNs. 
Before concluding, we present our experiments.

\section{Conditional Probabilistic Modeling}

In this paper, we are interested in predicting a collection of random variables $\Y$ given inputs $\X$, i.e.~we are interested in the problem of 
\emph{structured output prediction} (SOP).
In the probabilistic setting, SOP translates to modeling and learning a high-dimensional conditional distribution $P(\Y \cbar \X)$.
While one could learn a univariate predictor for each variable $Y \in \Y$ separately,
this approach assumes complete independence among $\Y$. This mean field assumption is often violated but still frequently used. 

Gaussian Processes (GPs) \citep{rasmussen2003gaussian} and Conditional Random Fields (CRFs) \citep{Lafferty:2001:CRF:645530.655813} are more expressive alternatives. 

However, they have serious shortcomings when inference has to scale to high-dimensional data or many samples.

One approach to scale GPs to larger datasets are deep mixtures of GPs, introduced in \citep{trapp2018learning}, which can be seen as a combination of GPs and SPNs.
However, while partially alleviating GP inference scalability issues, they are limited to continuous domains, while CSPNs can learn conditional distributions over heterogeneous data, i.e., where $\Y$ might contain discrete or continuous random variables, or be of mixed data types. In a nutshell, CSPNs can tackle \emph{unrestricted} SOP in a principled probabilistic way.

As an alternative within the family of tractable probabilistic models, logistic circuits (LCs) have been recently introduced as discriminative
models \citep{LiangAAAI19}, showing classification
accuracy competitive to neural nets on a series of
benchmarks. However, LCs and discriminative learning of
SPNs \citep{Gens2012} are limited to single output
prediction. Likewise, discriminative arithmetic circuits
(DACs) also directly tackle modeling a conditional distribution
\citep{rooshenas2016discriminative}. They are learned via
compilation of CRFs, requiring sophisticated and potentially
slow structure learning routines. Also related are  sum-product-quotient networks (SPQNs) \citep{sharir2017sum}, which extend SPNs by introducing quotient nodes. 
This enables SPQNs to represent a conditional distribution $P(\Y \cbar \X)$ as the ratio $P(\Y,\X) / P(\X)$ where the two terms are modeled by two SPNs. 
However, SPQNs do not incorporate neural networks, so that CSPNs can represent this ratio more compactly.

Our neural CSPNs are close in spirit to probabilistic models based on neural networks.
Generally, this line of research faces the challenge of how to parameterize
distributions using the outputs of deterministic neural function approximators.
Frequently, the mean field assumption is made, interpreting the output of the network
as the parameters of primitive univariate distributions, assuming independence
among random variables. Modelling complex distributions must then involve sampling
as in VAEs \citep{Kingma2014} or hierarchical variational models
\citep{ranganath2016hierarchical}, causing
significant computational overhead and yielding highly intractable models. Other approaches
for conditional density estimation based on neural networks include 
(conditional) normalizing flows, which yield tractable likelihoods, but are limited to continuous distributions and come with significant computational costs for computing the determinant of the Jacobian \citep{Rezende2015}.

Generally, CSPNs are most closely related to two classic approaches, namely mixture
density networks (MDNs) \citep{bishop1994mixture} and mixtures of experts (MoEs), in particular their hierarchical variant \citep{jordan1994hierarchical}. These models use the output of neural networks to parameterize a (typically Gaussian) mixture model. 
Shallow mixture models however are often inadequate in high dimensions, as the 
number of components required to accurately model the data may grow exponentially
with the number of dimensions \citep{Delalleau2011}. 
SPNs address this by instead encoding a hierarchy of mixtures, alternating between sums and products. Neural CSPNs may consequently be seen as a deep, hierarchical version of MDNs and MoEs.

\section{(Unconditional) Sum-Product Networks}

Before introducing our conditional sum-product networks (CSPNs), we briefly review classical (unconditional) SPNs.
For further details on SPNs, see
\citep{Darwiche2003,Poon2011,Peharz2017}. 
In the following, we denote random variables (RVs) as upper-case letters, e.g., $V$, their values as lower-case letters, e.g., $v\sim V$; and sets of RVs in bold, e.g., $\mathbf{v} \sim \V$.

A sum-product network (SPN) over a set of random variables $\V$ is defined via an acyclic directed graph (DAG) containing three types of nodes: distribution nodes, sum nodes and product nodes.
All leaves of the DAG are distribution nodes, and all internal nodes are either sums or products.
An SPN leaf represents a univariate distribution $P(Y)$ for some RV $Y \in \V$.
A sum node represents a \emph{mixture} $\sum_k w_k P_k(\Y)$, where $P_k$ are the children of the sum node according to the DAG, and where $w_k \geq 0$, and $\sum_k w_k = 1$.
We require that sum nodes are \emph{complete}, i.e.~that all children are defined over the same \emph{scope} $\Y$. 
A product node represents a \emph{factorization} $\prod_k P_k(\Y_k)$, where $P_k$ are the children of the product node according to the DAG.
We require that product nodes are \emph{decomposable}, i.e.~that all children are defined over \emph{disjoint scopes}, such that for any two product children $P_{i}(\Y_{i})$ and $P_{j}(\Y_{j})$, it holds that $\Y_{i} \cap \Y_{j} = \emptyset$. 
Completeness ensures that sum nodes $\sum_k w_k P_k(\Y)$ compute proper mixture models, and decomposability ensures that product nodes $\prod_k P_k(\Y_k)$ compute a factorized distribution assuming independence among the children.
Since the SPN leaves are distributions, it follows by induction that any node in an SPN represents a distribution over its scope.
SPNs are often assumed to have a single root with scope $\V$ that represents, by definition, the distribution modelled by the SPN.

In comparison to classical probabilistic graphical models, SPNs have the advantage that
they can represent certain distributions much more succinct, even if the corresponding
graphical model would have very high tree-width \citep{koller2009}.
Furthermore, \emph{inference routines} such as \emph{marginalization} and \emph{conditioning} can be tackled in linear time w.r.t. the network size \citep{Darwiche2003,Poon2011,Peharz2015b}.

These tasks can also be \emph{compiled}, i.e., starting from an SPN representing $P(\V)$, it is possible to generate a new SPN representing the marginal distribution for an arbitrary $\X \subset \V$, potentially conditioned on event $\Y = \mathbf{y}$, for any $\Y \subset \V$, $\Y \cap \X = \emptyset$.

Indeed, one way to represent conditional distribution $P(\Y \cbar \X)$ is to train a regular SPN
on the joint $P(\V)$, and to then compile it into a conditional distribution for each
input $\mathbf{x}$ of interest.
It is clear, however, that this will generally deliver sub-optimal results,
since the network will not be specialized to the specific input-output pair of $\X$ and $\Y$.
Intuitively, training the full joint $P(\V)$ optimizes \emph{all}
possible conditional distributions one can derive from the joint.
Thus, when we are interested in learning a conditional distribution for a set of
input and output variables $\X$ and $\Y$ known a-priori, directly learning
a conditional distribution, as discussed in the following, can be expected to deliver better results.

\section{Conditional Sum-Product Networks}

We now introduce a notion of conditional SPNs (CSPNs), which structurally reflect conditional independencies, and propose a learning framework to induce CSPNs from data.
To this end, we employ $\Y \subset \V$ to denote the target RVs, also called labels, while we denote the disjoint set of observed RVs, also called features, as $\X := \V \setminus \Y$.

\begin{figure}
\centering

\includegraphics[scale=0.5]{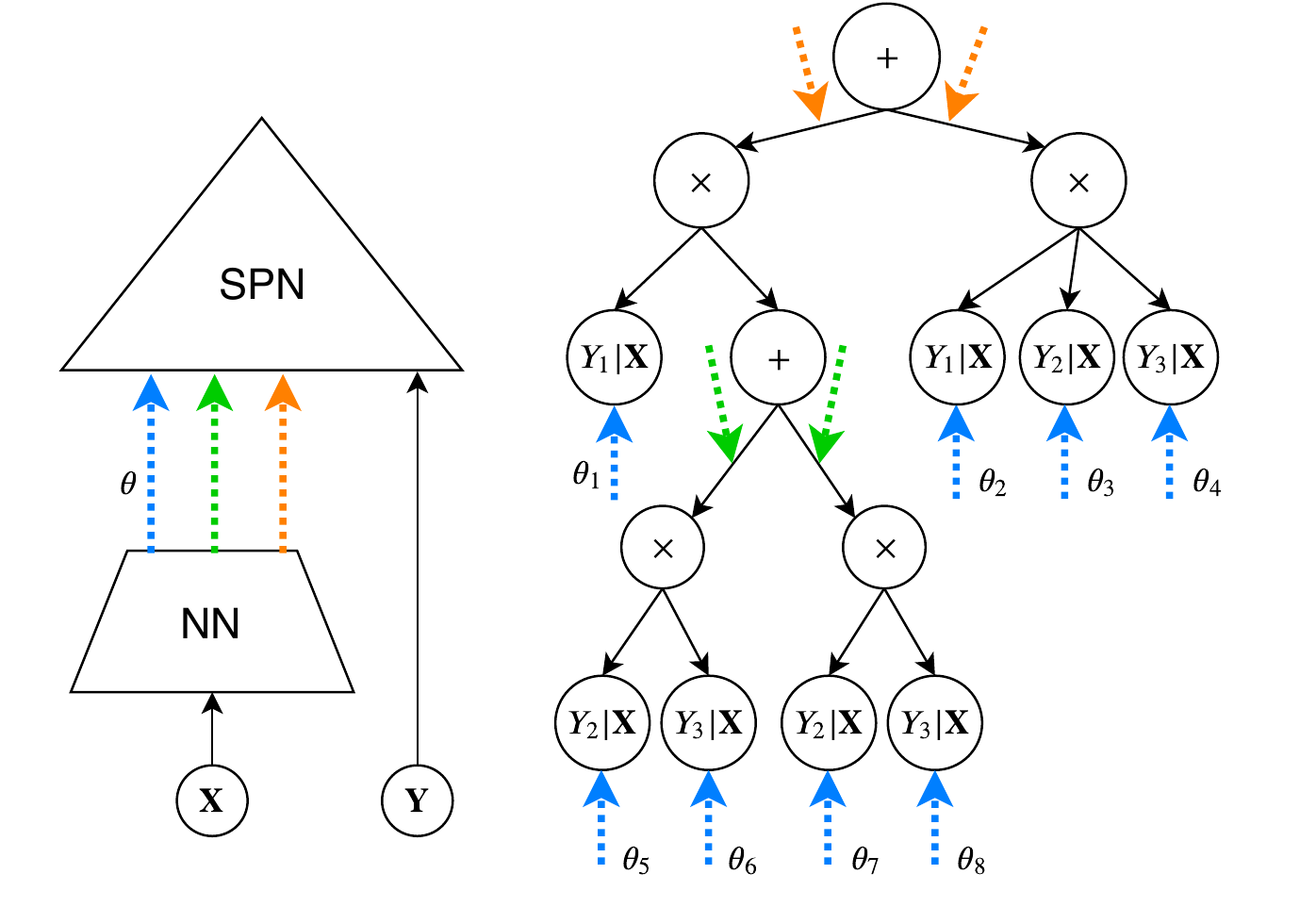}
\caption{Overview of the architecture (left) and a concrete CSPN example encoding $P(\Y \cbar \X)$ (right). $\X$ is the set of conditional variables and $\Y$ consists of three RVs $Y_{1}$, $Y_{2}$ and $Y_{3}$. Each color of the arrow represents one data flow. Here, the gating weights, possibly also leaf nodes, are parameterized by the output of a neural networks given $\X$.
}
\label{fig:cspnexample}
\end{figure}

{\bf Definition of Conditional SPNs (CSPNs).} 
We define a CSPN as a rooted DAG containing three types of nodes, namely
\emph{leaf}, \emph{gating}, and \emph{product} nodes, encoding a conditional probability distribution $P(\Y \cbar \X)$.
See Fig.~\ref{fig:cspnexample} for an illustrative example of a CSPN.
Each leaf encodes a normalized univariate conditional distribution $P(Y \cbar \X)$ over a target RV $Y \in \Y$, where $Y$ is denoted as the leaf's \textit{conditional scope}.

A product node factorizes a conditional probability distribution over its children, i.e., $\prod_{k}P_{k}(\Y_{k} \cbar  \X)$ where $\Y_{k} \subset \Y$.
To encode functional dependencies on the input features, a \emph{gating} node
computes $\sum_{k} g_k(\X)P_k(\Y \cbar \X)$ where $g_k$ is the output of a
nonnegative function $g$ w.r.t. the $k$-th child node, such that  $\sum_k g_k(\X) = 1$.
Standard sum nodes may still be realized by choosing a constant gating function.
The conditional scope of a non-leaf node is the union of the conditional scopes of its children.

Gating nodes are akin to gates in \textit{mixtures of experts}~\citep{shazeer2017outrageously}, which motives the name. The notions of completeness and decomposability of SPNs~\citep{Poon2011} naturally carry over to CSPNs,
and we can reuse the efficient SPN inference routines, guaranteeing that any conditional
marginal probability may be computed exactly in linear time~\citep{rooshenas2016discriminative}.

This can easily be verified by considering that for a fixed input $\x$, a CSPN reduces to a standard
SPN over the labels $\Y$, allowing any inference routine for standard SPNs
\citep{Poon2011,Peharz2015b,Vergari2018} to be employed.

{\bf CSPNs are More Compact than SPNs.}
It is known that neural networks are universal approximators, i.e., they can approximate any function to an arbitrary small error. As Choi and Darwiche (\citeyear{pmlr-v72-choi18a}) argue, a probabilistic model\footnote{In their exposition, they focus on Bayesian networks and arithmetic circuits, but their arguments carry over to SPNs as well.} represents not only one function, but a function for each possible probabilistic query.
Furthermore, as they argue, the functions corresponding to queries are, unlike neural networks, not universal approximators, as they are restricted to multi-linear functions of primitive distributions
or quotients thereof. While complex distributions may still be fit using a large number of e.g.\ 
Gaussian leaf distributions, this is wasteful if only a conditional distribution is required. 

The gating nodes of CSPNs extend SPNs in a way that allows them to also induce functions which are universal approximators.
For instance, using threshold gates $x_i\leq c$ ($c\in\mathbb{R}$), one can realize testing arithmetic circuits \citep{pmlr-v72-choi18a} which have been proven to be universal approximators. Within CSPNs we could for example also use single-output softmax gates in order to facilitate modelling mixtures of experts models in the form of Nguyen {\it et al.}~(\citeyear{Nguyen-mixtureofexperts2016}). This provides an alternative univeral approximation result, which shows that the class of CSPNs
mean functions is dense in the class of all continuous
functions over arbitrary compact domains.

\section{Learning Conditional SPNs}

A simple way to construct CSPNs representing $P(\Y \cbar \X)$ is to start from a standard SPN
over $\Y$, and to make its parameters (i.e., the sum weights and parameters of leaf
distributions) functional of input $\x$, defining $P(\Y \cbar \X=\x) = P(\Y ; \theta)$,
where parameters $\theta = f(\x)$ are a function of the input $\x$.
For $f$ we might use arbitrary function representations,
such as deep neural networks.
This architecture, which we call \emph{neural} CSPN, works reasonable well, as we demonstrate in the experimental section.
However, it fails to exploit conditional independencies in the structure. We introduce a structure learning strategy extending the established LearnSPN algorithm~\citep{Gens2013} which has been instantiated several times for
learning (unconditional) SPNs under different distributional assumptions~\citep{Vergari2015,Molina2018}.

Our \textit{LearnCSPN} routine builds a CSPN top-down by introducing nodes while partitioning
a data matrix whose rows represent samples and columns RVs in a recursive and greedy manner.
LearnCSPN is sketched in Alg.~\ref{algo:learncspn}.
It creates one of the three node types at each step: (1) a leaf, (2) a product, or (3) a gating node.
If only one target RV $Y$ is present, one conditional probability distribution
can be fit as a leaf.
To generate product nodes, conditional independencies are found by means of a
statistical test to partition the set of target RVs $\Y$.
If no such partitioning is found, then training samples are partitioned into clusters (conditioning) to induce a gating node.
We now review the three steps of LearnCSPN more in detail.

\begin{algorithm}[tb]
  \caption{\textsf{LearnCSPN} ($\mathcal{D}$, $\eta$, $\alpha$)}
  \label{algo:learncspn}
  \begin{algorithmic}[1]
    \REQUIRE data $\mathcal{D}=\{(\mathbf{y}_i, \mathbf{x}_i) | \mathbf{y}_i\in \Y, \mathbf{x}_i\in \X\}_{i=1}^{N}$ where $\mathbf{y}_i =  \left ( y_i^1, \cdots, y_i^{c_{y}} \right )$, and $\mathbf{x}_i =  \left ( x_i^1, \cdots, x_i^{c_{x}} \right )$; $\eta$: min
    number of instances to split; $\alpha$: threshold of significance
    \ENSURE a CSPN $S$ encoding $P(\Y \cbar \X)$ learned from $\mathcal{D}$
    \IF {$|\Y| = 1$}
    \STATE $S\leftarrow \mathsf{LearnConditionalLeaf}(\mathcal{D})$ using, e.g., GLMs 
    \ELSIF {$|\mathcal{D}| < \eta$}
    \STATE $S \leftarrow \prod\nolimits_{j=1}^{|\Y|}\mathsf{LearnCSPN}(\{(y_i^j, \mathbf{x}_i)\}_{i=1}^{N}, \eta, \alpha)$
    \ELSE
    \STATE $\{\V_{k}\}_{k=1}^K\leftarrow \mathsf{SplitLabels}(\mathcal{D}, \alpha)$ \textcolor{gray}{// compare Alg.~\ref{algo:splitFeatures}}
    \IF {$K > 1$}
    \STATE $\mathcal{D}_{k} \leftarrow \{\mathbf{v}_{k}^{m}|\mathbf{v}_k^{m}\sim \V_k\}_{m=1}^{M}$
    \STATE $S\leftarrow  \prod_{k=1}^{K}\mathsf{LearnCSPN}(\mathcal{D}_{k}, \eta)$
    \ELSE
    \STATE $\{\mathcal{D}_{k}\}_{k=1}^K\leftarrow \mathsf{SplitInstances}(\mathcal{D})$ e.g. using random splits or k-Means with an appropriate metric
    \STATE $S\leftarrow \sum_{k=1}^{K} 
    g_{k}(x)\cdot \mathsf{LearnCSPN}(\mathcal{D}_{k}, \eta)$, $x \in D$ where $g_{k}(x)$ is a gating function
    \ENDIF
    \ENDIF
    \RETURN $S$
  \end{algorithmic}
\end{algorithm}

\begin{algorithm}[tb]
  \caption{\textsf{SplitLabels} ($\mathcal{D}$, $\alpha$)}
  \label{algo:splitFeatures}
  \begin{algorithmic}[1]
    \REQUIRE data $\mathcal{D}=\{(\mathbf{y}_i, \mathbf{x}_i)|\mathbf{y}_i\in \Y, \mathbf{x}_i\in \X\}_{i=1}^{N}$ where the label RVs are $\Y=\{Y_{1},\dots,Y_{P}\}$, $\alpha$: threshold of significance 
    \ENSURE a label partitioning $\{\mathcal{P}_{\mathcal{D}}\}$
    \STATE $\mathcal{G}\leftarrow\mathsf{Graph(\{\})}$ 
    \FOR{{\bfseries each} $Y_{i}, Y_{j}\in \Y$}
        \STATE $S_{i, j}\leftarrow M \| \hat{\sum}_{Y_{i} Y_{j}\cdot \X} \|_{F }^{2}$
        \IF {$\mathsf{LindsayPillaBasak}(S_{i,j}) > \alpha$}
            \STATE $\mathcal{G}\leftarrow\mathcal{G}\cup\{(i, j)\}$
        \ENDIF
    \ENDFOR
  \RETURN $\mathsf{ConnectedComponents}(\mathcal{G})$
  \end{algorithmic}
\end{algorithm}

\textbf{(1) Learning Leaves.}
In order to allow for tractable inference, we require conditional models at the leaves to be normalized.
Apart from this requirement, \textit{any} such univariate tractable conditional model may be
plugged in a CSPN effortlessly to model $P(Y \cbar \X)$. This can be simple univariate models
or the joint output of a deep model. 

In this paper, we use Generalized Linear Models (GLMs) \citep{mccullagh1984generalized} as leaves, due to their simple and flexible nature.
We compute $P(y \cbar \mu=\glm(\X))$ by regressing univariate parameters $\mu$ from features $\X$, for a given set of distributions in the exponential family.

\textbf{(2) Learning Product Nodes.}
For product nodes, we are interested in decomposing the labels $\Y$ into subsets via conditional independence (CI). In terms of density functions, testing that RVs $Y_i$ are independent of $Y_j$ given $\X=\mathbf{x}$, for any value of $\mathbf{x}$, i.e., $Y_i \bigCI Y_j | \mathbf{X}$, can equivalently be characterized as $p(Y_i,Y_j \cbar \X) = p(Y_i \cbar \X)p(Y_j \cbar \X)$.

Since we aim to accommodate arbitrary leaf conditional distributions in CSPNs, regardless of their parametric likelihood models or data types (i.e.~discrete or continuous), we adopt a non-parametric pairwise CI test procedure to decompose labels $\Y$.
Kernel-based methods like KCIT~\citep{zhang2012kernel} and PCIT~\citep{doran2014permutation} would be an adequate and powerful choice, but scale unfortunately quadratically in the sample size.
For this reason, we employ a randomized approximation of KCIT, the randomized conditional correlation test (RCoT) \citep{strobl2017approximate}, which has been shown to deliver very similar results as KCIT and PCIT, while scaling only linearly in the sample size.

Briefly, RCoT computes the same statistics as KCIT, i.e., the squared Hilbert-Schmidt norm of the partial cross-covariance operator but uses the Lindsay-Pilla-Basak method to approximate the asymptotic distribution.
In order to do that, RCoT specifies conditional independence using characteristic kernels (e.g. RBFs, Laplacian) $k_{\mathcal{Y}_i}, k_{\mathcal{Y}_j}, k_\mathcal{X}$ for variables $Y_i,Y_j, \X$ from domains $\mathcal{Y}_i, \mathcal{Y}_j, \mathcal{X}$ and denote their corresponding RKHS by $\mathcal{H}_{\mathcal{Y}_i}, \mathcal{H}_{\mathcal{Y}_j}, \mathcal{H}_{\mathcal{X}}$. Then it employs the cross-covariance operator on the RKHS from $\mathcal{H}_{\mathcal{Y}_i}$ to $\mathcal{H}_{\mathcal{Y}_j}$ which is defined as $\langle f, \sum \nolimits_{Y_iY_j}g \rangle = \mathbb{E}_{Y_i Y_j}[f(Y_i)g(Y_j)] - \mathbb{E}_{Y_i}[f(Y_i)]\mathbb{E}_{Y_j}[g(Y_j)]$
for all $f \in \mathcal{H}_{\mathcal{Y}_i}$ and $g \in \mathcal{H}_{\mathcal{Y}_j}$. The partial cross-covariance operator $\sum \nolimits_{Y_i Y_j \cdot \X} $ of $(Y_i,Y_j)$ given $\X$ can then be written as $\sum \nolimits_{Y_i Y_j\cdot \X} = \sum \nolimits_{Y_i Y_j} - \sum \nolimits_{Y_i \X}\sum \nolimits_{\X \X}^{-1}\sum \nolimits_{\X Y_j}$. Under mild assumptions, it then holds:
if $Y_i  \bigCI Y_j | \X$ 
then $\mathbb{E}_{\X}[P({Y_i Y_j  \cbar  \X})] = \mathbb{E}_{\X}[P(Y_i \cbar \X)P({Y_j \cbar \X})]$ 
and in turn
$\sum \nolimits_{Y_i Y_j\cdot \X} = 0\;$ \footnote{Indeed, there are some special cases where $Y_i \centernot{\bigCI} Y_j | \X$ yet $\sum \nolimits_{Y_i Y_j\cdot \X} = 0$, i.e., this is not an equivalence relation. However, these cases are rarely encountered in practice.} . Refer to \citep{strobl2017approximate} for further details.

In the end, we create a graph where the nodes are RVs in $\Y$ and put an edge between two nodes $Y_i, Y_j$ if we cannot reject the null hypothesis that $Y_i, Y_j\bigCI \X$ for a given threshold $\alpha$.
The conditional scopes of product children are then given by connected components of this graph, akin to \citep{Gens2013}, cf.~ Alg.~\ref{algo:splitFeatures}.

\textbf{(3) Learning Gating Nodes.}
Gating nodes provide a mechanism to condition on $\X$, enhancing flexibility.
We first select a functional form for the gating function $g_k(\X)$, ideally, a differentiable parametric one such as logistic regression or a neural network. This function is restricted to allow for a proper mixture of distributions, i.e., $\sum_k g_k(\X) = 1$ and $\forall_\X g_k(\X) >= 0$.

For structure learning, we perform clustering over features $\X$, and denote the corresponding member assignment as a one-hot coded vector $\Z$. 
To this end, one can exploit any clustering scheme based on the available knowledge of the data
distribution (e.g., k-Means for Gaussians). We can also leverage random splits, as in random projection
trees \citep{dasgupta2008random}.

We then proceed to fit the gating function to predict $\Z_k = g_k(\X)$. This is a generalization over sum nodes, where $g_k(\X) = c$ that is, the mixtures are constants independent of $\X$.
This functional mixing does not break the tractability guarantees over $\Y$ as $\X$ is always assumed to be evidence in the conditional case.

\textbf{End-to-End Parameter Optimization.}
The CSPNs described here contain two sets of parameters: one for the weights of the gating nodes, and one for the parameters of the leaf distributions. \textsf{LearnCSPN} in Alg.~\ref{algo:learncspn} automatically sets the parameters of the gating function as described above.
The parameters of the GLMs are obtained by an Iteratively Reweighted Least Squares (IRWLS)
algorithm as described in \citep{green1984iteratively}, on the instances available at the
\emph{leaf} node. However, those parameters are locally optimized and usually not optimal
for the global distribution. Fortunately, CSPNs are differentiable as long as the
leaf models and gating functions are differentiable. Hence, one can optimize the
conditional likelihood in an end-to-end fashion using gradient based optimization
techniques. We follow this approach for training neural CSPNs.

\section{ABCSPN: Autoregressive Block-wise CSPN}

To illustrate how to impose structure on generative models by employing CSPNs as
building blocks, in the same way as Bayesian networks represent a joint distribution
as a factorization of conditional models, we build an autoregressive model composed of CSPNs.
By applying the chain rule of probabilities, we can decompose a joint distribution as the product $P(\Y,\X)=P(\Y \cbar \X)P(\X)$.
Then, one could learn an SPN to model $P(\X)$ 
and a CSPN for $P(\Y \cbar \X)$.
By combining both models using a single product node, we represent the whole joint as a computational graph. 

Now, if one applies the same operation several times by repeatedly partitioning $\Y$ in a series of disjoint sets $\Y_{1},\Y_{2},\ldots$ we can obtain an \textit{autoregressive} model representation.
Inspired by image autoregressive models like PixelCNN~\citep{van2016conditional}, and PixelRNN \citep{vandenOord2016} we propose an
\textit{Autoregressive Block-wise CSPN} (ABCSPN) for conditional image generation. 
For one ABCSPN, we divide images into pixel blocks, hence factorizing the joint distribution block-wise instead of pixel-wise as in PixelC/RNN. 
Each factor accounting for a block of pixels is then a CSPN representing the  distribution of those pixels as conditioned on all previous blocks and on the class labels.\footnote{Here, image labels play the role of observed RVs in $\X$.}

We factorize blocks in raster scan order, row by row and left to right, but any other ordering is also possible.
The complete generative model over image $\mathbf{I}$ encodes:
$p(\textbf{I}) = \prod\nolimits_{i=1}^{n} p(\mathbf{B}_{i} \cbar \mathbf{B}_{1}, \dots, \mathbf{B}_{i-1}, \mathbf{C}) \cdot p(\mathbf{C})$
where $\textbf{B}_{i}$ denotes the pixel RVs of the $i$-th block and $\mathbf{C}$ the one-hot coded image class. 
Learning each conditional block as a CSPN can be done by the structure learning routines introduced above.

%
%
%

\begin{figure}[!t]
    \centering
    \includegraphics[scale=0.28]{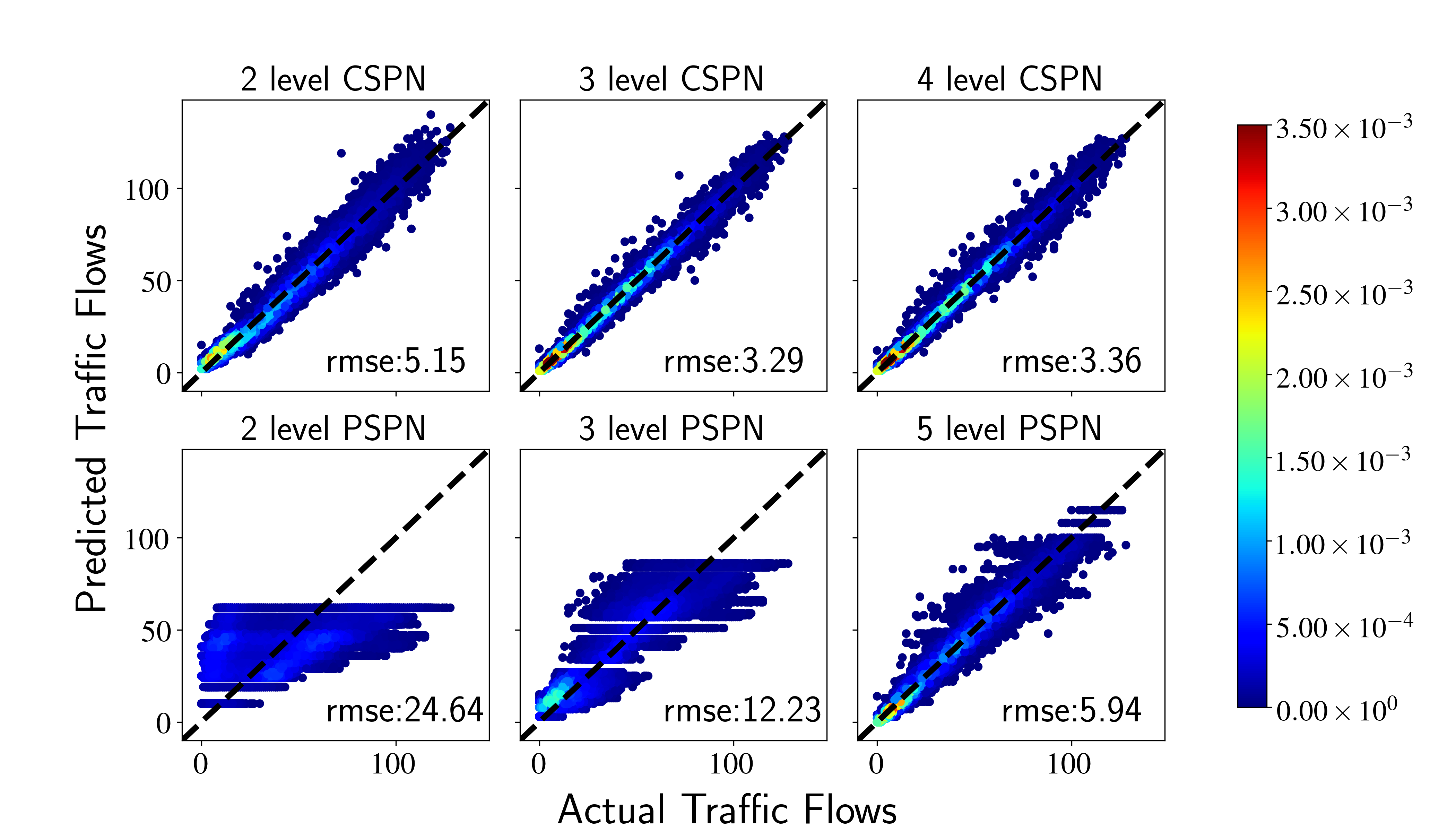}
    \caption{Comparing traffic flow predictions (RMSE, the lower the better) of Poisspn CSPNs (top) versus SPNs (botton, PSPNs) for shallow (left) or deep models (center and right).
    CSPNs are consistently more accurate than corresponding SPNs and, as expected, deeper CSPNs outperform shallow ones (center and right). (Best viewed in color)}
\label{fig:cologn}
\end{figure}

\section{Experimental Evaluation}
Here we investigate CSPNs in experiments on real-world data. 
Specifically, we aim to answer the following questions: 
\textbf{(Q1)} Can CSPNs perform better than regular SPNs?  
\textbf{(Q2)} How accurate are CSPNs for SOP?
\textbf{(Q3)} How do ABC-SPNs perform w.r.t. state-of-the-art generative models? 
\textbf{(Q4)} Do neural CSPNs outperform baseline neural conditional models such as MDNs on image SOP tasks?

To this end, we implemented CSPNs in Python calling TensorFlow and R.


\textbf{(Q1, Q2) Multivariate Traffic Data Prediction.}
We employ CSPNs for multivariate traffic data prediction, comparing them against SPNs with Poisson leaf distributions~\citep{Molina2017}. This is an appropriate model as the traffic data represents counts of vehicles.
We considered temporal vehicular traffic flows in the German city of Cologne \citep{ide2015lte}. The data comprises 39 RVs whose values are from stationary detectors located at the 50km long Cologne orbital freeway, each one counting the number of vehicles within a fixed time interval.

It contains 1440 samples, each of which is a snapshot of the traffic flow.
The task of the experiments is to predict the next snapshot ($|\Y|=39$) given a historical one ($|\X|=39$). 

We trained both CSPNs and SPNs controlling the depth of the models. The CSPNs use GLMs with exponential link function as parameter for a Poisson univariate conditional leaf. Results are summarized in Fig.~\ref{fig:cologn}.
We can see that CSPNs are always the most accurate model as their root mean squared error (RMSE) is always the lowest.
As expected, deeper CSPNs have lower predictive error compared to shallow CSPNs. Moreover smaller CSPNs perform equally well or even better than SPNs, empirically confirming that CSPN are more expressive than SPNs.
This answers \textbf{(Q1, Q2)} affirmatively and  also provides evidence for the convenience of directly modeling a conditional distribution.

\begin{table}[t]
\caption{Average test conditional log-likelihood (CLL) of DACL and CSPN on $20$ standard density estimation benchmarks. Significantly better results (t-test,$p<0.05$) are bold. As the number of wins, draws and losses (W/D/L) of CSPNs shows, CSPNs are competitive to DACL.}\smallskip

\centering
\begin{tabular}{lllll}
\toprule
             & \multicolumn{2}{l}{ \thead{50\% Evidence} } & \multicolumn{2}{l}{\thead{80\% Evidence}} \\
\midrule
Dataset      & DACL                 & CSPN                 & DACL                 & CSPN                 \\
\midrule
Nltcs        &   -2.770               &  -2.795       &     -1.255        &    -1.256           \\ 
Msnbc        & \textbf{-2.918}        &  -3.165       &  \textbf{-1.557}          &    -1.684             \\
KDD  &  -0.998       &    -1.023            &  -0.386             &  -0.397             \\
Plants       & -4.655             &    -4.720            & -1.812            & \textbf{-1.683}              \\
Audio        &  -18.958            &   -18.543     & -7.337            &       \textbf{-7.110}       \\
Jester       &  -24.830         &     -24.543          &  -9.998          &   \textbf{-9.830}             \\
Netflix      & -26.245           &     \textbf{-25.914}          &  -10.482        & -10.351                \\
Accidents    &   \textbf{-9.718}            &    -11.587          &   \textbf{-3.493} &  -4.045             \\
Retail       &  \textbf{-4.825}             &    -5.600          &   -1.687            &        -1.654             \\
Pumsb.   &  \textbf{-6.363}             &     -7.383         &  -2.594            &   -2.618            \\
Dna          & \textbf{-34.737}             & -38.243          &   -12.116             &      -11.895          \\

Kosarek      &    \textbf{-5.053}                  &    -5.527                  &               -2.549       &       \textbf{-2.397}               \\
MSWeb        &    \textbf{-5.653}                  &    -6.686                  &                -1.333      &       -1.335               \\
Book         &    -16.801                  &   \textbf{-10.653}                   &             -6.817         &         \textbf{-3.191}            \\
EachMovie    &     -25.325                 &   \textbf{-18.130}                   &              -9.403        &       \textbf{-4.579}               \\
WebKB        &     -72.072                 &   \textbf{-18.542}                   &             -28.087         &       \textbf{-2.623}               \\
Reuters-52   &    -41.544                  &   \textbf{-15.736}                   &             -17.143         &       \textbf{-3.878}               \\
20News &      -76.063                &    \textbf{-35.900}                  &                   -27.918   &             \textbf{-4.984}         \\
Bbc          &    -118.684                  &   \textbf{-47.138}                   &            -44.811          &      \textbf{-2.996}                \\
Ad           &    \textbf{-4.893}                  &   -6.290                   &               -1.370       &      -1.030               \\
\midrule
\textbf{W/D/L CSPN} &  \multicolumn{2}{c}{\textbf{ 7/5/8 }}  &   \multicolumn{2}{c}{\textbf{ 10/8/2} }\\
\bottomrule
\end{tabular}
\label{tab:mlc_nll}
\end{table}

\textbf{(Q2) Conditional Density Estimation.}
We now focus on conditional density estimation. Due to space constraints, we present results on a subset of the standard binary benchmark datasets\footnote{We adopted the data splits of~\citeauthor{rooshenas2016discriminative} (\citeyear{rooshenas2016discriminative}).}, when different percentage of evidence ($|\X|$) is available. 
We compare to DACL~\citep{rooshenas2016discriminative} as it currently provides state-of-the-art conditional log-likelihoods (CLLs) on such data.
To this end, we first perform structure learning on the train data split (stopping learning when no more than 10\% of samples are available), followed by end-to-end learning on the train and validation data. Note that the sophisticated structure learning in DACL directly optimizes for the CLL at each iteration.

Tab.~\ref{tab:mlc_nll} reports statistically significant results (best in bold), after a paired t-tests ($p < 0.05$) has been run. 
We can see how on the 80\% evidence scenario CSPNs are comparable with DACL on most benchmarks.
On the other hand, in case only 50\% of $\X$ is observable, DACL tends to perform better than CSPNs, even though by a slight margin in general. We note that CSPNs are faster to learn than DACL and that, in practice, no real hyperparameter tuning was necessary to achieve these scores, while DACL ones are the result of a fine grained grid search (see~\citep{rooshenas2016discriminative}. This answers \textbf{(Q2)} affirmatively and shows that CSPNs are comparable to state-of-the-art.

\textbf{(Q3) Auto-Regressive Image Generation.}
We investigate ABCSPNs on a subset (20000 random samples) of grayscale MNIST and Olivetti faces by splitting each image into 16 resp. 64 blocks of equal size where we normalized the greyscale value for MNIST. Then we trained a CSPN on Gaussian domain for each block conditioned on all the blocks above and to the left of it and on the image class and formulate the distribution of the images as the product of all the CSPNs. We compare with the state-of-the-art PixelCNN++ model \citep{salimans2017pixelcnn++}. 
Training PixelCNN++ on a machine with 4 NVIDIA GeForce GTX 1080 GPUs took approximately a week to converge, while ABCSPNs employ one order of magnitude less parameters and converge in hours.
Additionally, PixelCNN++ generates images pixel by pixel in sequence, whereas ABCSPNs yields a block at a time. That is, our structure speeds up also the generative process.

\begin{figure}[!t]
\centering
\includegraphics[scale=0.4,width=0.06\columnwidth]{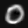}\hspace{-2pt}
\includegraphics[scale=0.4,width=0.06\columnwidth]{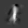}\hspace{-2pt}
\includegraphics[scale=0.4,width=0.06\columnwidth]{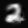}\hspace{-2pt}
\includegraphics[scale=0.4,width=0.06\columnwidth]{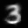}\hspace{-2pt}
\includegraphics[scale=0.4,width=0.06\columnwidth]{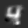}\hspace{-2pt}
\includegraphics[scale=0.4,width=0.06\columnwidth]{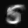}\hspace{-2pt}
\includegraphics[scale=0.4,width=0.06\columnwidth]{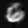}\hspace{-2pt}
\includegraphics[scale=0.4,width=0.06\columnwidth]{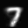}\hspace{-2pt}
\includegraphics[scale=0.4,width=0.06\columnwidth]{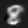}\hspace{-2pt}
\includegraphics[scale=0.4,width=0.06\columnwidth]{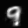}\\
\includegraphics[scale=0.4,width=0.06\columnwidth]{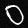}\hspace{-2pt}
\includegraphics[scale=0.4,width=0.06\columnwidth]{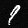}\hspace{-2pt}
\includegraphics[scale=0.4,width=0.06\columnwidth]{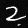}\hspace{-2pt}
\includegraphics[scale=0.4,width=0.06\columnwidth]{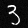}\hspace{-2pt}
\includegraphics[scale=0.4,width=0.06\columnwidth]{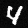}\hspace{-2pt}
\includegraphics[scale=0.4,width=0.06\columnwidth]{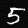}\hspace{-2pt}
\includegraphics[scale=0.4,width=0.06\columnwidth]{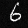}\hspace{-2pt}
\includegraphics[scale=0.4,width=0.06\columnwidth]{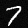}\hspace{-2pt}
\includegraphics[scale=0.4,width=0.06\columnwidth]{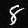}\hspace{-2pt}
\includegraphics[scale=0.4,width=0.06\columnwidth]{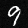}\hspace{-2pt}
\caption{Samples generated by an ABCSPN (top), and PixelCNN++ (bottom) trained on MNIST.\label{fig:pixelmnist}}
\end{figure}

As can be seen Figs.~\ref{fig:pixelmnist} and \ref{fig:pixelolivetti}, samples from ABCSPNs look as plausible as PixelCNN ones. Furthermore, ABCSPNs achieves this while reducing the number of independency tests among pixels required by CSPNs: from quadratic over all pixels in an image down to quadratic in the block size.

\begin{figure}[!t]
    \centering
    \hspace*{-1.2cm} 
    \includegraphics[scale=0.07]{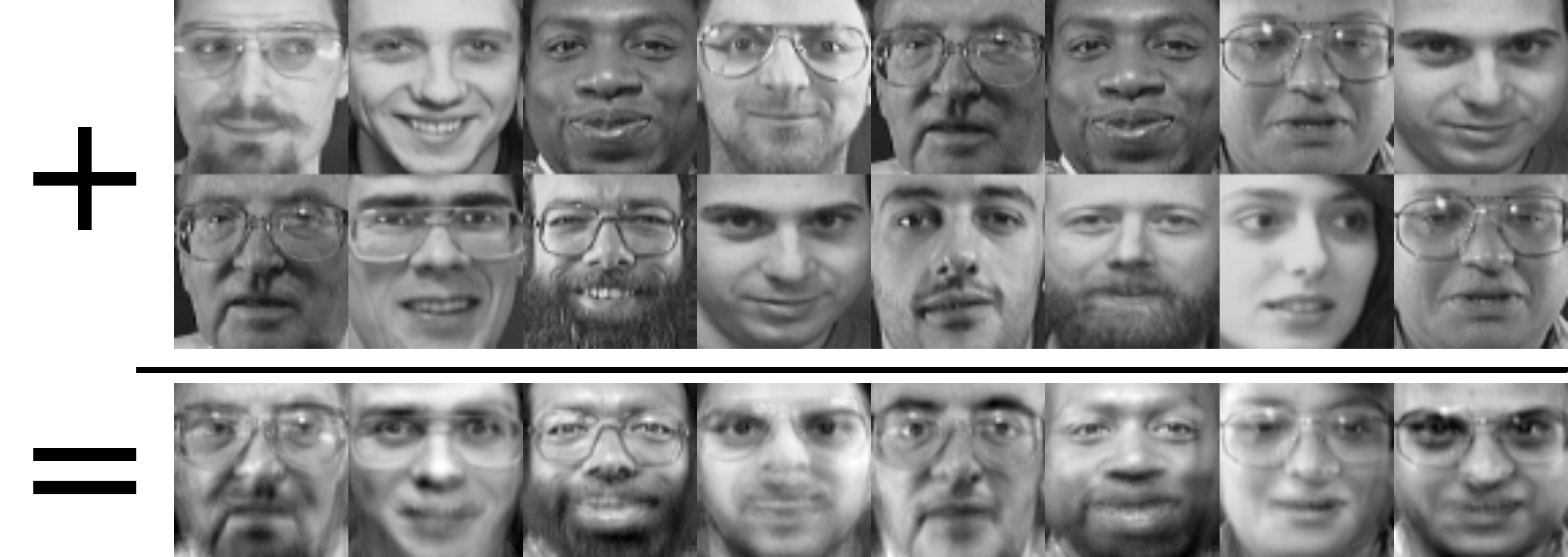}
    \caption{Conditional image generation with ABCSPNs: bottom row images are sampled while conditioning on the two classes to which individuals from the two upper rows belong.}
    \label{fig:interpolation}
\end{figure}

In Fig.~\ref{fig:interpolation}, new faces are sampled from an ABCSPN after conditioning on a set of class images that is the mixing of two original classes in the Olivetti dataset.
That is, by conditioning on multiple classes it generates samples that resemble both individuals belonging to those classes, even though the ABCSPN never saw that class combination before during training.
This demonstrates how ABCSPNs are able to learn meaningful and accurate models over the image manifold, providing an affirmative answer to \textbf{(Q3)}.

\textbf{(Q4) Neural CSPNs with Random Structures.}
In high-dimensional domains, such as images, the structure learning procedure introduced above may be intractable. 
In these cases, CSPNs may still be applied by starting from a random SPN structure
as proposed by Peharz et al.\ \citeyear{Peharz2019}, resulting in a flexible distribution
$P(\Y ; \theta=f(\X))$.
When $f$ is represented by a deep neural network, we obtain a highly
expressive conditional density estimator which can be trained end-to-end.

To demonstrate the efficacy of this approach, we evaluate it on several multilabel image classification tasks.
The goal of each task is to predict the joint conditional distribution of binary labels $Y$ given
an image $X$. Experiments were conducted on the CelebA dataset, which features images of faces annotated with
40 binary attributes. In addition, we constructed multilabel
versions of the MNIST and Fashion-MNIST datasets, by adding additional
labels indicating symmetry, size, etc.\ to
the existing class labels, yielding 16 binary labels total.

We compare our model to two different common ways of parameterizing conditional
distributions using neural networks.
The first is the mean field approximation, whereby the output of a neural network is
interpreted as logits of independent univariate Bernoulli distributions, assuming
that the labels $Y$ are conditionally independent given $X$.
Second, we compare to mixture density networks with 10 mixture components, each itself a
mean field distribution.
For each of these models, including the CSPN, we use the same standard convolutional neural network
architecture up to the last two layers. Those final two layers are customized to the
different desired output formats: For the mean field and MDN models, all parameters
are predicted using two dense layers. For the CSPN, we instead use a dense layer followed
by a 1d-convolution, in order to obtain the increased number of SPN parameters
without using drastically more neural network weights.
The resulting conditional log-likelihoods as well as accuracies are given in
Table \ref{tab:mlc_neural}. To compute accuracies, we obtained MPE estimates from the
models using the standard max-product approximation.
On the MNIST and Fashion dataset, estimates were counted as accurate only if all 16 labels
were correct, on the CelebA dataset, we report the average accuracy across all 40 labels.
The results indicate that the commonly used mean field
approximation is inappropriate on the considered datasets, as allowing the inclusion
of conditional dependencies resulted in a pronounced increase in both likelihood and accuracy.
In addition, the improved model capacity of the CSPN compared to the MDN yielded a further
performance increase. On CelebA, our CSPN outperforms a number of
sophisticated neural network architectures from the literature,
despite being based on a standard convnet with only about 400k parameters \citep{ehrlich2016facial}.

\begin{table}[tb]
\caption{Average test conditional log-likelihood (CLL) and test accuracy of the 
mean field (MF) model, mixture density network (MDN), and neural conditional
SPN (CSPN) on multilabel image classification tasks.
Predictions on MNIST and Fashion are counted as accurate only if all 16 labels are correct. For
CelebA, we report the average accuracy across all labels. The best results are marked in bold. As one can see, the additional representational power of CSPNs yields notable improvements.} \smallskip
\centering
\begin{sc}
\begin{tabular}{lrrrrrr}
\toprule
& \multicolumn{3}{c}{CLL}& \multicolumn{3}{c}{Accuracy}\\
\midrule
            & MF       & MDN      & CSPN    & MF   & MDN  & CSPN \\
\midrule
MNIST       &   -0.70  &  -0.61   & \bf -0.54  & 74.1\% & 76.4\% & \bf 78.4\% \\ 
Fashion     &   -0.95  &  -0.73   & \bf -0.70  & 73.4\% & 73.7\% & \bf 75.5\% \\ 
CelebA      &   -12.1 &  -11.6 &   \bf -10.8 & 86.6\%  & 85.3\% & \bf 87.8\% \\ 
\bottomrule
\end{tabular}
\end{sc}
\label{tab:mlc_neural}
\end{table}

\section{Conclusions}
We have extended the concept of sum-product networks (SPNs) towards conditional distributions by 
introducing conditional SPNs (CSPNs). Conceptually, they combine simpler models in a hierarchical fashion in order to create a deep representation that can model multivariate and mixed conditional distributions while maintaining tractability. They can be used to impose structure on deep probabilistic  models and, in turn, significantly boost their power as demonstrated by our experimental results. 

Much remains to be explored, including other learning methods for
CSPNs, design principles for CSPN+SPN architectures, combining the (C)SPN stack with the deep neural learning stack, more work on extensions to
sequential and autoregressive domains, and further applications.

\section{Acknowledgements}
We acknowledge the support of the German Science Foundation (DFG) project "Argumentative Machine Learning” (CAML, KE 1686/3-1) of the SPP 1999 “Robust Argumentation Machines” (RATIO). Kristian Kersting also acknowledges the support of the Rhine-Main Universities Network for “Deep Continuous-Discrete Machine Learning” (DeCoDeML). 
Thomas Liebig was supported by the German Science Foundation under project B4 `Analysis and Communication for Dynamic Traffic Prognosis' of the Collaborative Research Centre SFB 876.

\bibliographystyle{named}
\bibliography{refs}
\end{document}